\documentclass[times,twocolumn,final]{elsarticle}

\usepackage{style}
\usepackage{framed,multirow}

\usepackage{amssymb}
\usepackage{latexsym}

\usepackage{url}
\usepackage{xcolor}

\usepackage[colorlinks=true,linkcolor=black,citecolor=black,urlcolor=blue]{hyperref}

\usepackage{booktabs}
\usepackage{amsmath}
\usepackage{soul}
\usepackage{adjustbox}
\usepackage{subcaption}  
\usepackage{float}       

\definecolor{newcolor}{rgb}{.8,.349,.1}

\renewenvironment{abstract}{\global\setbox\absbox=\vbox\bgroup
  \hsize=\textwidth%
  \noindent\unskip\ignorespaces}
 {\egroup}

\begin{document}

\begin{frontmatter}

\title{Monocular pose estimation of articulated open surgery tools - in the wild\tnoteref{aam}}
\tnotetext[aam]{Accepted for publication in Medical Image Analysis. This is the author accepted manuscript (AAM) made available under the CC BY-NC-ND 4.0 license (\url{https://creativecommons.org/licenses/by-nc-nd/4.0/}). The Version of Record is available at \url{https://doi.org/10.1016/j.media.2025.103618}.}

\author[1]{Robert \snm{Spektor}\corref{cor1}}
\ead{spektor@campus.technion.ac.il}

\author[2]{Tom \snm{Friedman}}

\author[2]{Itay \snm{Or}}

\author[2]{Gil \snm{Bolotin}}

\author[1]{Shlomi \snm{Laufer}}

\address[1]{Faculty of Data and Decision Sciences, Technion - Israel Institute of Technology, Haifa, Israel}
\address[2]{Rambam Health Care Campus, Haifa, Israel}

\begin{abstract}
This work presents a framework for monocular 6D pose estimation of surgical instruments in open surgery, addressing challenges such as object articulations, specularity, occlusions, and synthetic-to-real domain adaptation. The proposed approach consists of three main components: $(1)$ synthetic data generation pipeline that incorporates 3D scanning of surgical tools with articulation rigging and physically-based rendering; $(2)$ a tailored pose estimation framework combining tool detection with pose and articulation estimation; and $(3)$ a training strategy on synthetic and real unannotated video data, employing domain adaptation with automatically generated pseudo-labels.
Evaluations conducted on real data of open surgery demonstrate the good performance and real-world applicability of the proposed framework, highlighting its potential for integration into medical augmented reality and robotic systems. The approach eliminates the need for extensive manual annotation of real surgical data.
\end{abstract}


\end{frontmatter}


\section{Introduction}
\label{sec1}
Object pose estimation is a fundamental problem in computer vision that aims to determine the position and orientation of an object in 3D space relative to a camera. It has a wide range of applications, including robotic manipulation \cite{tremblay2018deep}, autonomous navigation \cite{mousavian20173d}, and augmented reality \cite{su2019deep}. In the medical domain, pose estimation of surgical instruments plays a crucial role in computer-assisted surgery, skill assessment, and robotic assistance \cite{burton2023evaluation,bkheet2023using,doignon2008pose}.

Existing approaches to object pose estimation can be broadly categorized into two groups: marker-based \cite{meza2021markerpose, ababsa2004robust} and markerless methods. Marker-based methods rely on attaching physical markers, such as color-coded fiducials \cite{wei1997real} or passive reflective markers \cite{elfring2010assessment}, to the instruments. These markers are then detected and tracked in the image space to estimate the pose. While marker-based methods can provide accurate pose estimates, they require modifying the instruments and may interfere with the surgical workflow \cite{bouget2015detecting}.

On the other hand, markerless methods aim to estimate the pose directly from the visual appearance of the instruments in the image. These methods typically involve training deep neural networks on large datasets of annotated images. The networks learn to extract relevant features and estimate the 6D pose of the instruments using techniques such as correspondence estimation with Perspective-n-Point (PnP) algorithm \cite{su2022zebrapose, wang2021gdr}. Markerless methods offer a more practical and non-intrusive solution for surgical instrument pose estimation, as they do not require any modifications to the instruments.

While markerless pose estimation methods have shown promising results, they face several challenges that hinder their widespread adoption in surgical settings. One of the primary challenges is the scarcity of annotated real-world data \cite{allan20192017, bouget2017vision}. Collecting and annotating large datasets of images with accurate 6D pose labels is time-consuming, expensive, and often impractical \cite{hodan2018bop, tremblay2018deep}. This challenge is particularly relevant in the medical domain, where privacy concerns and the difficulty of recording during surgeries due to camera placement restrictions (as some areas must remain sterile) further complicate the data collection process. 

To address the data scarcity problem, researchers have explored the use of synthetic data generation techniques \cite{tremblay2018training, tremblay2018deep}. By creating realistic virtual models of surgical instruments and rendering them in simulated environments, large amounts of annotated training data can be generated efficiently. However, models trained solely on synthetic data often suffer from the domain gap problem, where the learned features do not generalize well to real-world scenarios. To bridge this gap, domain adaptation techniques, such as domain randomization \cite{tremblay2018deep} and adversarial training \cite{bai2021recent}, have been proposed. These techniques aim to make the models more robust to the differences between synthetic and real data, improving their performance on real-world images.

\begin{figure}[!t]
  \centering
  \includegraphics[scale=0.34]{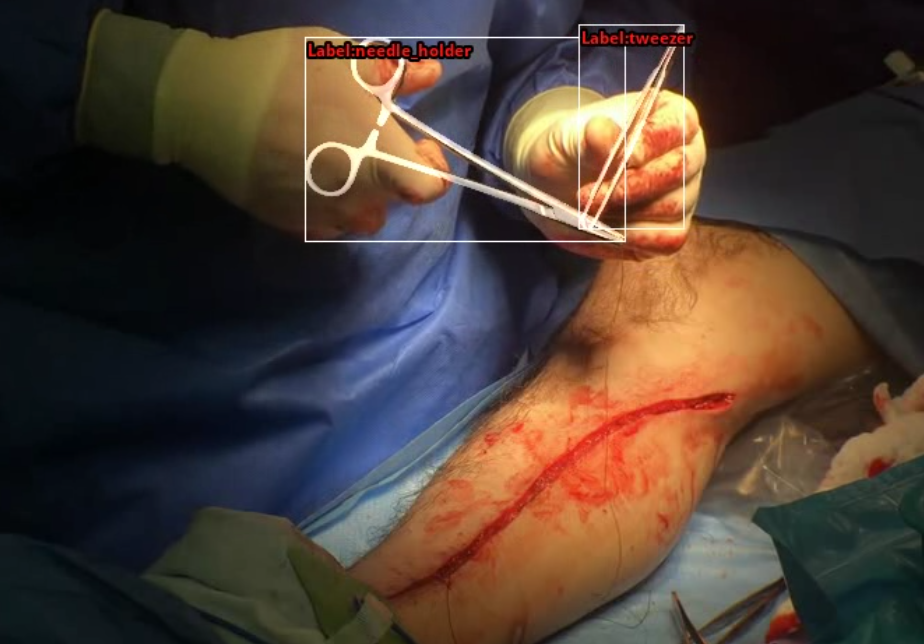}
  \caption{Pose estimation of surgical instruments, showcasing the precision required in labeling articulated tools, such as needle holders and tweezers, during an open surgical procedure.}
  \label{challenge}
\end{figure}

Despite the significant progress in deep learning-based pose estimation methods, their application to surgical instruments poses additional challenges due to the unique characteristics of the surgical environment and the instruments themselves \cite{nema2022surgical, colleoni2020synthetic}. Surgical instruments often have metallic and reflective surfaces, which can cause severe specular reflections and make it difficult to extract reliable visual features. Moreover, articulated objects, such as surgical instruments with movable parts, introduces additional complexities. Articulated objects have additional degrees of freedom due to their moving parts, which significantly increases the difficulty of accurately estimating their pose.

\begin{figure}[!t]
  \centering
  \includegraphics[scale=.52, angle=270]{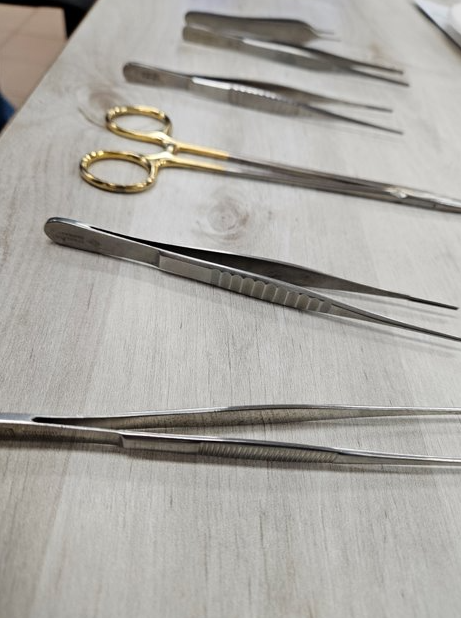}
  \caption{Surgical instruments with reflective surface}
  \label{reflective}
\end{figure}

Our key contributions include:
\begin{enumerate}
\item Creation of a new in-the-wild dataset of open surgery tool and hand segmentations.
\item Development of a synthetic dataset for tool pose estimation that addresses issues like tool articulation and hand occlusions.
\item Pose estimation network designed for articulated surgical tools without predefined geometric assumptions.
\item Introduction of a synthetic-to-real domain adaptation technique that leverages the temporal features of video data to enhance both object detection and pose estimation models, enabling mutual improvement between the two.
\end{enumerate}
It is important to emphasize that our approach removes the requirement for manual data labeling. New tools can be incorporated into the dataset using only their 3D models (e.g., STL files), without the need for manual annotation. Any labeling performed in this study is strictly for the purpose of validating our algorithms. 

\section*{Related Work}
\label{Related}
\subsection{Object Pose Estimation Methods}
The advent of deep learning has propelled significant advancements in object pose estimation, giving rise to diverse methodologies. These approaches can be broadly categorized into single-stage versus two-stage methods, direct pose regression techniques, and keypoint detection followed by PnP-RANSAC. These categories are not mutually exclusive, and many state-of-the-art methods combine elements from different approaches to achieve better performance.

Single-stage methods \cite{xiang2017posecnn, braun2016pose, he2017mask, zakharov2019dpod}, integrate object detection and pose estimation within a unified network, offering a streamlined process. This approach allows for end-to-end training and potentially faster inference times. In contrast, two-stage methods \cite{li2019cdpn, park2019pix2pose, peng2019pvnet, su2022zebrapose, wang2021gdr}, decouple the object detection and pose estimation tasks. While this separation may increase computational complexity, it allows for independent optimization of each stage, potentially leading to improved accuracy.

Keypoint detection methods can be further categorized into sparse and dense techniques \cite{peng2019pvnet, wang2021gdr, su2022zebrapose}. Sparse keypoint detection identifies a limited number of significant points on the object within the image. Dense keypoint detection recognizes a more extensive set of points across the object's surface in the image, potentially providing richer information for pose estimation. For instance, \cite{peng2019pvnet} employs a voting-based method to handle occluded or truncated keypoints, demonstrating the robustness of keypoint-based approaches.

After keypoint identification, pose estimation techniques generally fall into two categories: Classical PnP-RANSAC \cite{peng2019pvnet, su2022zebrapose} and direct pose regression \cite{wang2021gdr}. The classical PnP-RANSAC approach uses the identified keypoints to solve the Perspective-n-Point (PnP) problem, often combined with RANSAC for robust estimation. Direct regression techniques aim to predict pose parameters in an end-to-end fashion.

Recent works have introduced innovative approaches. \cite{su2022zebrapose} focuses on dense surface encoding by assigning unique codes to 3D object model vertices and then uses a PnP-RANSAC variant to determine the pose. \cite{wang2021gdr} utilizes dense intermediate geometric representations and employs a direct regression method to estimate the pose.

While these methods have shown impressive results on general object pose estimation benchmarks, they often face challenges when applied to surgical instruments due to the unique characteristics of the surgical environment. In response to these challenges, researchers have developed specialized methods for surgical instrument pose estimation. For laparoscopic instruments, \cite{hasan2021detection} proposed a two-stage approach. Their method employs a CNN-based segmentation network for instrument detection and segmentation, followed by a separate CNN for 3D pose estimation using algebraic geometry. This approach demonstrated high accuracy on a dataset of laparoscopic images.
For open surgery scenarios, \cite{hein2021towards} introduced a markerless hand-tool tracking pipeline, focusing on larger, non-articulated tools. Their approach involved creating synthetic data using the MANO hand model \cite{MANO:SIGGRAPHASIA:2017} and GraspIt! simulator \cite{miller2004graspit}, capturing real-world data in a mock operating room for fine-tuning, and evaluating multiple baseline models. 
In a subsequent multi-view approach \cite{hein2023next} introduced a multi-camera capture setup (including both static and head-mounted cameras) and a multi-view RGB-D dataset of ex-vivo spine surgeries, evaluating single- and multi-view pose estimation methods for surgical instrument tracking. Their results demonstrated sub-millimeter accuracy for certain instruments under optimal conditions, suggesting that markerless, multi-view tracking can be a feasible alternative to existing marker-based systems.

Synthetic data generation has gained significant attention in the object pose estimation community. The common approach involves 3D modeling and rendering, using physically-based rendering (PBR) techniques to create realistic images with accurate pose annotations. The effectiveness of this approach was demonstrated in the BOP challenge 2020, where methods trained on PBR images significantly outperformed those trained on simpler "render \& paste" images.

While synthetic data generation has shown great promise in training deep learning models for object pose estimation, there often exists a domain gap between synthetic and real-world data \cite{tremblay2018deep}. Various domain adapatation techniques have been proposed to bridge this domain gap and improve the transferability of models trained on synthetic data to real-world applications.

Several works have successfully demonstrated the effectiveness of bridging the domain gap between synthetic and real data for object pose estimation. For instance, the work by \cite{hein2021towards} on hand-tool interaction in open surgery utilizes a combination of synthetic pretraining and real-world fine-tuning. They first train their models on a large-scale synthetic dataset generated using a physically-based rendering pipeline and then fine-tune the models on a smaller dataset of real-world images captured in a mock operating room. This approach allowed them to leverage the diversity and accuracy of synthetic data while adapting the models to the characteristics of real-world surgical scenes.

Similarly, \cite{souipas2023simps} achieve markerless tool segmentation and 3D localization without requiring prior knowledge of tool structure. Although they focused on articulated surgical tools, they did not explicitly handle the articulations and conducted their experiments on mockup operations. 

Another example is the work by \cite{sundermeyer2020augmented}, where they propose a self-supervised domain adaptation method for object pose estimation. Their approach relies on a differentiable renderer to generate synthetic views of objects and align them with real-world images using a contrastive loss. By minimizing the domain discrepancy between synthetic and real-world images, their method can improve the performance of object pose estimation models on real-world data without requiring any labeled real-world examples.

\section{Data Collection and Annotation}
For this work, we collected real-world data from surgical procedures for the purpose of unsupervised sim-to-real domain adaptation and to validate our approach and ensure its effectiveness in practical surgical scenarios. 

\subsection{Data Collection}
We collected data from ten different surgeries. Each surgery involved the removal of a vein from the leg as part of the "Coronary Artery Bypass Graft (CABG)" procedure. Our focus is specifically on the suturing part of the procedure, which typically lasts between 15 to 30 minutes. In total, five different surgeons performed the surgeries, with each suturing procedure carried out by a single surgeon. The study was approved by the Rambam Health Care Campus IRB committee.

\subsection{Data Annotation for Evaluation}
\label{sec:data_annot}
To ensure unbiased evaluation, three surgical videos were reserved exclusively for domain adaptation and the other seven for testing. Approximately 100 images were selected from each testing procedure for evaluation, covering a variety of scenarios.

Since annotating 6D pose is very complex, we instead annotate instance segmentation labels and evaluate on them. For each selected frame, we created detailed segmentation masks of the surgical tools of interest (needle-holder and tweezers), as well as background surgical tools and hands. These segmentation masks provide pixel-level annotations of the tools and hands in the surgical scene. The annotation process was carried out using the SAM (Segment Anything) model \cite{kirillov2023segany}, which generates high-quality object masks from input prompts such as points or boxes. Any masks that did not meet quality standards were manually adjusted. These annotated frames provide essential ground truth data for evaluating the effectiveness of our pose estimation network on actual surgical data.

\section{Methodology}
\subsection{Overview of the Proposed Pipeline}

The proposed framework addresses the unique challenges of the surgical environment, such as the scarcity of annotated real-world data, the geometry of articulated instruments, and the presence of occlusions.

The core goal is to enable accurate pose estimation without extensive manual annotation of real surgical data. To achieve this, a three-component pipeline is proposed, consisting of synthetic data generation, a tailored pose estimation network, and a training strategy that leverages both synthetic and real unannotated data.

\subsection{Synthetic Data Generation}
\label{sec:synth}

\begin{figure}[!t]
  \centering
  \includegraphics[scale=0.635]{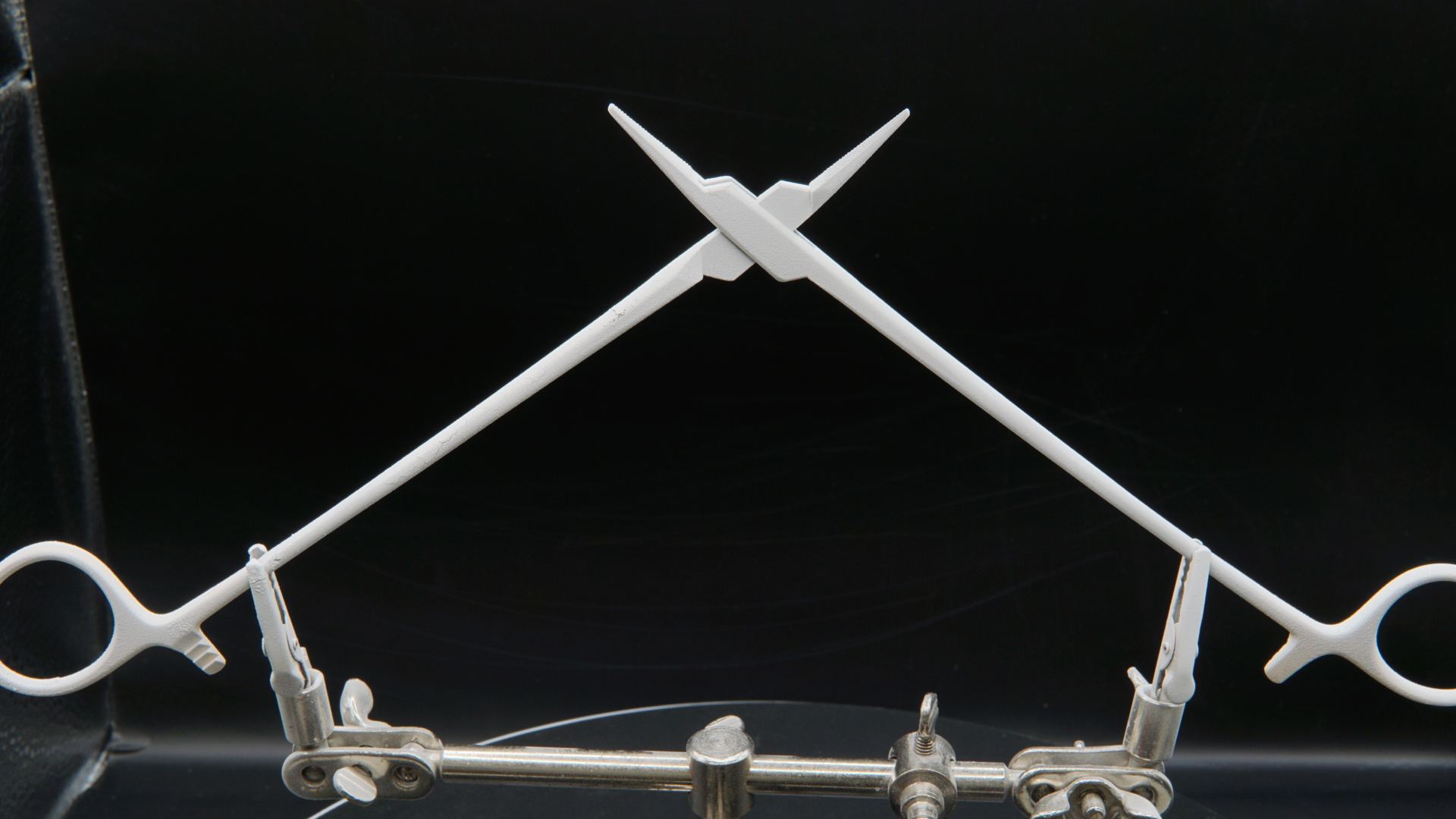}
  \caption{Needle-holder photographed on a spinning table for photogrammetry.}
  \label{challenge}
\end{figure}

Synthetic data helps overcome the challenges associated with acquiring and annotating real-world data, which is particularly difficult in the surgical domain due to privacy concerns and the complexity of capturing ground truth poses.

\subsubsection{3D Modeling of Surgical Instruments}
We developed custom CAD models for our surgical instruments due to a lack of suitable existing options and the need for highly realistic and accurate tool representations. To achieve this level of detail and accuracy, we employed photogrammetry techniques to capture the intricate features of the actual instruments.

To prepare the surgical instruments for photogrammetry, we spray them with a dust spray to minimize glare and reflections that could interfere with the scanning process. Dry shampoo, surprisingly, outperformed specialized 3D scanning sprays, providing better reconstructions at a lower cost. The tools are then placed on a spinning table and photographed 360 degrees at three different heights. This approach ensures that every detail and surface of the instruments is accurately captured, providing us with the high-quality data needed to create our custom CAD models.

The captured images are processed using Reality Capture \cite{RealityCapture}, a professional photogrammetry software. Once the initial 3D meshes are generated, we perform a series of refinement steps to enhance their quality and realism. We use Blender \cite{BlenderFoundation2023}, a popular free and open source 3D modeling and animation software, to clean the meshes by removing noise and artifacts introduced during the photogrammetry process. Additionally, we employ Blender's smoothing and hole-filling tools to improve the mesh surface quality.

To simulate the articulation of the surgical instruments, we used two different techniques. For the needle holder, which consists of two rigid parts, we separated the model into two parts. For the tweezers, which are made from a single part, the opening and closing introduce bending. To simulate this bending, we used Blender's rigging and weight-painting tools. Rigging involves creating a virtual skeleton within the 3D mesh, allowing us to define the movement of different parts. Weight painting assigns influence levels to the bones, which control how the mesh deforms when the bones move. This combination allows us to simulate the bending motion of the tweezers, ensuring realistic articulation.

\begin{figure}[!t]
  \centering
  \includegraphics[scale=.12]{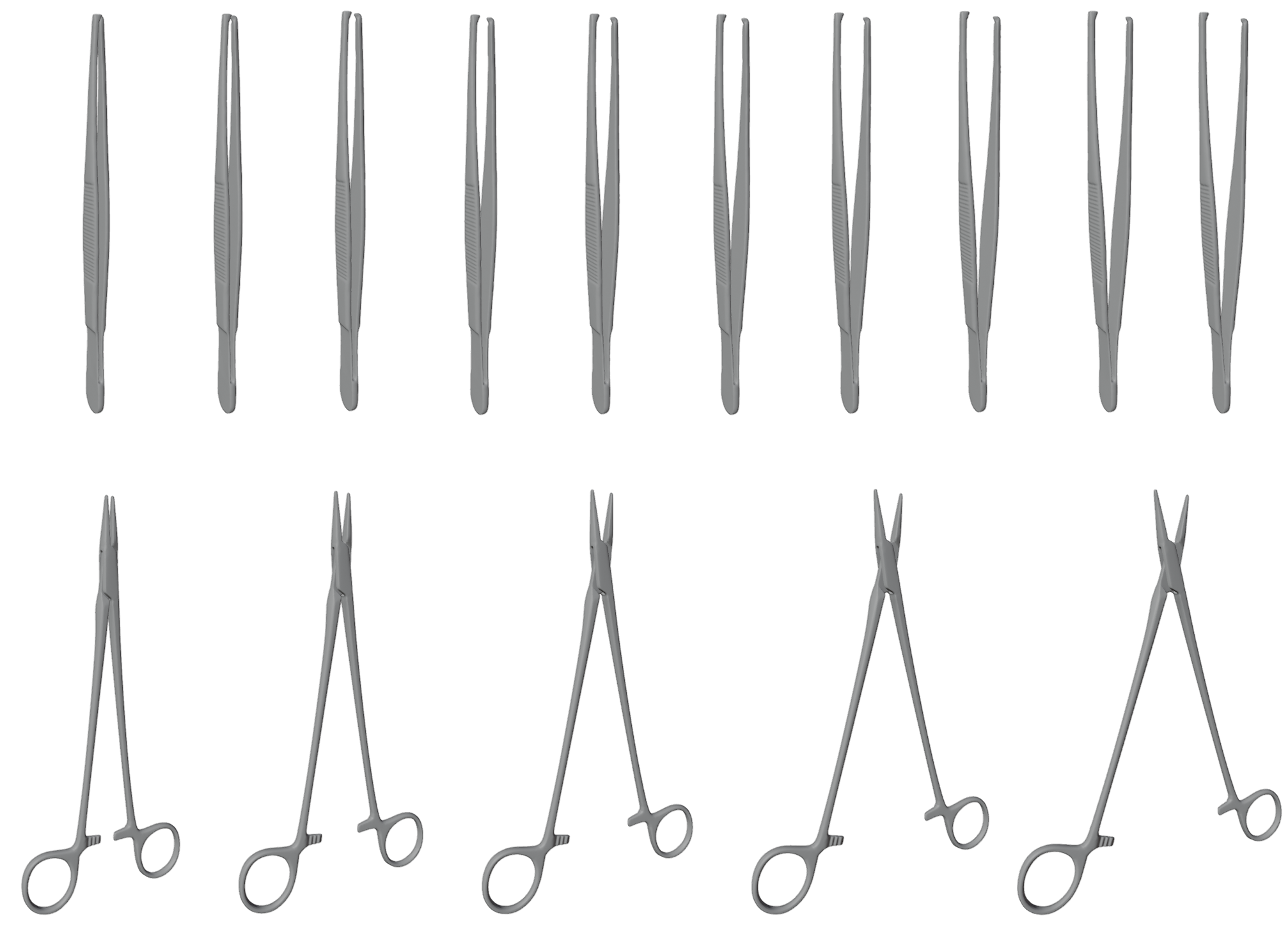}
  \caption{Generated synthetic surgical tools in varying degrees of articulation}
  \label{tool_articulation}
\end{figure}

Figure \ref{tool_articulation} shows the tools we scanned at different articulation angles. After rigging the meshes, we exported them in different articulation angles. Specifically, we exported 15 different meshes for the needle holder and 10 meshes for the tweezers, each representing a different articulation angle.

\subsubsection{Hand-Object Interaction Modeling}
\label{sec:hand_object_inter}
Occlusions of hand grasps are commonly encountered in surgical scenarios. We simulate them with ContactGen \cite{liu2023contactgen}, a generative hand-grasp model, in combination with the MANO hand mesh \cite{MANO:SIGGRAPHASIA:2017}. We used the model from the official implementation, which was trained on the GRAB dataset \cite{GRAB:2020, Brahmbhatt_2019_CVPR}. The GRAB dataset includes real human grasps for 51 objects from 10 different subjects.

We generate a diverse set of hand-object grasps by varying the grip positions, hand orientations, and contact points between the hand and the surgical instruments. By incorporating these hand-object grasps into our synthetic data, we introduce realistic hand occlusions and interactions.

\subsubsection{Rendering Pipeline}
With the rigged 3D models of the surgical instruments and the hand-object grasps, we proceed to render a large synthetic dataset using Blender. Our rendering pipeline generates diverse and realistic images with varying backgrounds and lighting conditions.

We create virtual environments within Blender, including operating room scenes and other indoor environments. To enhance the realism and variability of the rendered images, we incorporate HDRI (High Dynamic Range Imaging) maps, which provide high-quality lighting and reflections. These HDRI maps cover a wide range of lighting scenarios, from surgical settings to other indoor environments, adding diversity to our synthetic dataset.

During the rendering process, we randomize various aspects of the scene to increase the diversity of the synthetic dataset. This includes randomizing the positions and orientations of the surgical instruments, the hand-object grasps, and the camera viewpoints. We also apply random variations to the material properties of the instruments, such as their color and surface properties, to introduce visual variability.

\begin{figure}[!t]
  \centering
  \includegraphics[scale=.57]{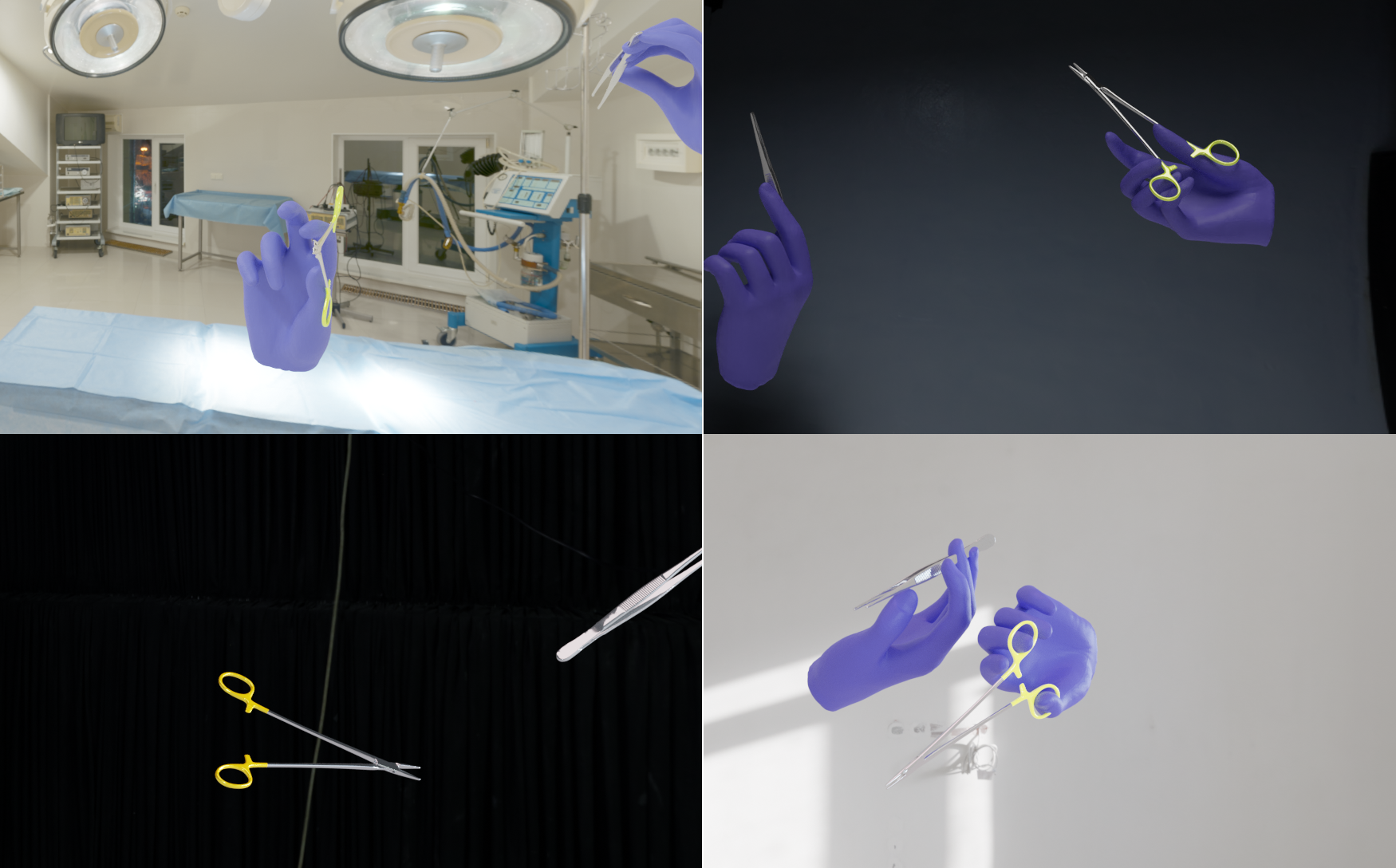}
  \caption{Synthetic data of surgical tools}
  \label{synth}
\end{figure}

In total, we generate 80,000 synthetic images using our rendering pipeline, utilizing BlenderProc \cite{Denninger2023}. Each image is accompanied by precise ground-truth annotations, including the 6D pose of the surgical instruments and the articulation angles. This comprehensive synthetic dataset serves as a valuable resource for training our pose estimation network. A sample from this dataset is illustrated in Figure \ref{synth}.

\subsection{Pose Estimation Framework}
\label{framework}

Our pose estimation framework combines object detection and 6D pose estimation stages with a domain adaptation strategy designed to bridge the gap between synthetic and real-world surgical data. This approach combines established, state-of-the-art components with novel adaptations tailored for the unique challenges of surgical tool pose estimation.

For the initial object detection stage, we employ YOLOv8 \cite{Jocher_Ultralytics_YOLO_2023}, a state-of-the-art object detection framework recognized for its speed and accuracy. We adapt its standard training process to incorporate our synthetic-to-real domain adaptation strategy, which is detailed further in Section \ref{training_procedure}.

The subsequent pose estimation stage receives cropped object regions identified by the detector. It then processes them using our pose estimation network. This network builds upon the strong foundation of GDR-Net \cite{wang2021gdr} and its enhanced implementation, GDRNPP \cite{liu2022gdrnpp_bop}, known for state-of-the-art performance in general 6D object pose estimation. From these baseline models, we adopt several key components proven effective for pose estimation: we utilize the continuous 6D rotation parameterization proposed by \cite{zhou2019continuity} for stable orientation representation, incorporate the differentiable Patch-PnP module \cite{wang2021gdr} to enable end-to-end pose regression during training, and employ multi-task loss functions for comprehensive optimization of pose-related objectives.

For domain adaptation, we integrate pseudo-labeling cycles inspired by iterative refinement strategies in prior work (e.g., \cite{hein2021towards}). This iterative process is employed during training to progressively refine model predictions on unlabeled real-world data, thereby mitigating the performance gap caused by the domain shift from the annotated synthetic source data.

\begin{figure*}[!t]
\centering
\includegraphics[width=\textwidth]{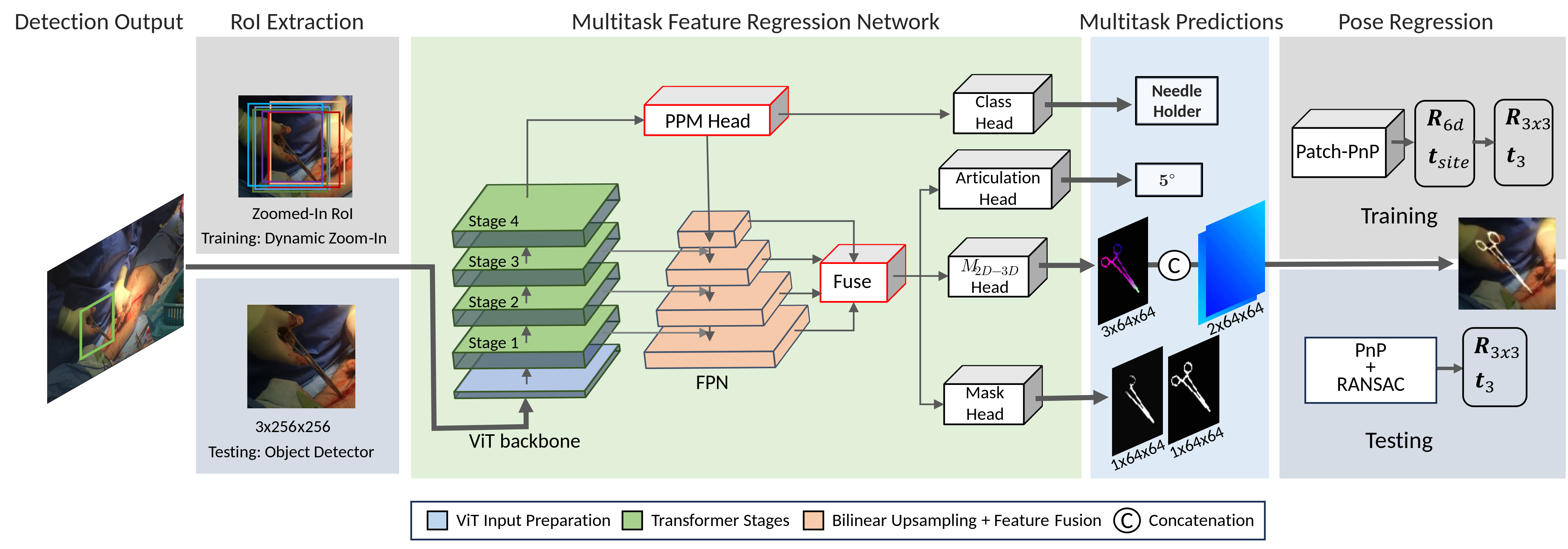}
\caption{
Pose Estimation Model Framework: An input RGB image is first cropped to the Region of Interest (RoI) and processed by the Multitask Feature Regression Network, which uses a Vision Transformer (ViT) backbone for feature extraction. The extracted features pass through a UPerNet architecture with a Pyramid Pooling Module (PPM) to aggregate multiscale contextual information, then branch into four output heads: object category classification, articulation angle regression, object masks, and 2D-3D dense correspondence mapping. The object masks help stabilize training and facilitate domain adaptation. The resulting dense correspondence output ($M_{2D-3D}$) is concatenated with CoordConv layers and then sent either to the Patch-PnP module (during training) or to PnP-RANSAC (during inference) for pose estimation.}

\label{fig:pose_model}
\end{figure*}

Our novel contributions include architectural and methodological adaptations. Architecturally, we introduce a multitask feature regression module (Figure \ref{fig:pose_model}) for simultaneous pose, category, and articulation angle estimation from image features. We also incorporate mechanisms to handle tool articulation and type variability, enabling generalization to new tools without requiring specific geometric models. Methodologically, our domain adaptation strategy extends pseudo-labeling techniques to simultaneously improve detection and pose estimation for articulated tools, using only synthetic data as the annotated source.

\subsubsection{Direct 6D Object Pose Estimation}
\label{sec:direc_pose}

As briefly mentioned in \ref{Related}, our pose estimation method draws inspiration from GDR-Net \cite{wang2021gdr} for direct 6D object pose estimation, specifically the GDRNPP implementation \cite{liu2022gdrnpp_bop}, which incorporates several improvements over the original GDR-Net presented in the conference version. It is worth noting that GDRNPP won most awards in the 2022 BOP challenge \cite{sundermeyer2023bop} for ``6D localization of seen objects" and ``2D detection of seen objects", demonstrating its state-of-the-art performance. We adopt several key components from GDR-Net, including the 3D rotation parameterization, 3D translation parameterization, 2D-3D dense correspondence maps, Patch-PnP, and the multitask 6D pose loss, which are described in the following paragraphs.

\paragraph{3D Rotation Parameterization} For the 3D rotation, we adopt the 6D rotation representation \cite{zhou2019continuity} instead of commonly used representations, such as quaternions or Euler angles. Quaternions and Euler angles suffer from ambiguities and discontinuities, which can hinder the learning process and lead to suboptimal pose estimates. The 6D rotation representation provides a continuous and unambiguous way of representing rotations.

The 6D rotation representation consists of two 3D vectors, denoted $r_1$ and $r_2$, which correspond to the first two columns of the rotation matrix $\mathbf{R}{\in}\textit{SO}(3)$. The network predicts these two columns using the Patch-PnP module. To ensure the orthogonality and validity of the rotation matrix, the Gram-Schmidt process is applied to the predicted columns. This process orthogonalizes the columns and computes the third column to complete the rotation matrix. The resulting rotation matrix $R = [r_1, r_2, r_3]$ is guaranteed to be orthogonal and represents a valid 3D rotation.

In the context of working with object crops, we focus on predicting the allocentric representation of an object's rotation \cite{kundu20183d}. This approach is independent of the viewer's perspective and assumes that the camera has rotated from its original position to point toward the center of the object's Region of Interest (RoI). This representation eliminates the need to account for variable camera-object positioning, simplifying the rotational estimation process. Once the allocentric rotation is determined, it can be converted into the true 3D rotation (egocentric representation) by using the object's 3D translation and the camera's intrinsic parameters $K$ \cite{kundu20183d}.

\paragraph{3D Translation Parameterization} Direct prediction of translation from a Region of Interest (RoI) is impossible without additional information. To address this, we use a parameterization technique known as the scale-invariant translation estimate (SITE) \cite{li2019cdpn}. SITE represents the translation as a combination of the 2D center coordinates of the object in the image plane and the depth of the object relative to the camera. The network predicts three components: $(\delta_x, \delta_y, \delta_z)$.

$\delta_x$ and $\delta_y$ represent the offset of the object's center with respect to the center of the bounding box in the image plane, normalized by the bounding size $(w,h)$. and $\delta_z$ represents the depth of the object relative to the camera, normalized by the bounding box zoom-in ratio $r=s_\text{zoom}/\text{max}(w,h)$ where $s_\text{zoom}$ is the crop zoom-in size.

To obtain the final 3D translation $t = (t_x, t_y, t_z)$, we first compute the 2D center coordinates of the object $(o_x, o_y)$ by adding the predicted offsets $(\delta_x, \delta_y)$ to the center of the bounding box $(c_x, c_y)$:
\begin{align*}
o_x &= c_x + \delta_x \cdot w \\
o_y &= c_y + \delta_y \cdot h
\end{align*}

Then, using the camera intrinsic parameters (focal length $f$ and principal point $(p_x, p_y)$), we compute the 3D translation:
\begin{align*}
t_x &= (o_x - p_x) \cdot \frac{t_z}{f} \\
t_y &= (o_y - p_y) \cdot \frac{t_z}{f} \\
t_z &= \delta_z \cdot r
\end{align*}

\paragraph{2D-3D Dense Correspondence Maps} The network generates a dense 2D-3D correspondence map \cite{li2019cdpn, hodan2020epos}, which is an intermediate geometric feature that establishes a relationship between each object pixel in the 2D image and its corresponding 3D coordinate on the object model. These 3D coordinates are normalized using the dimensions of a tight bounding box enclosing the object’s mesh. This normalized coordinate representation is referred to as NOCS maps \cite{wang2019normalized}. Generating these 2D–3D correspondences for articulated objects can be done in two ways: (1) fixing correspondences per articulation angle via its own bounding box, or (2) normalizing by the bounding box of the fully articulated configuration. We adopt the second approach to avoid maintaining a lookup table. We then use CoordConv layers \cite{liu2018intriguing} by concatenating (x, y) coordinates to the dense correspondences as channels, providing explicit 2D positional encoding. These layers are particularly useful after cropping the images. The dense correspondence map can then be used with PnP-RANSAC to solve for the object's 6D pose.

\begin{figure}[!t]
\centering
\includegraphics[scale=.54]{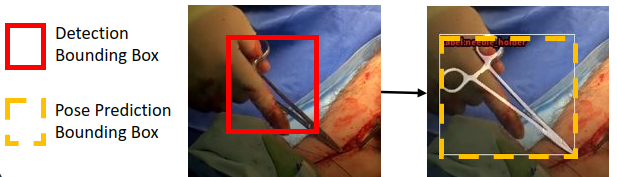}
\caption{Bounding box refinement stage: This figure illustrates the process of refining bounding boxes for object detection. Initially, frames with high detection confidence and consecutive tracking detections are selected. These frames are then passed to the pose estimator, which estimates the object pose, class, and articulation angle. The resulting mesh and pose are reprojected onto the image plane, creating a reprojected bounding box that is used as the new pseudo-label for retraining the object detection network.}
\label{fig:bbox_ref}
\end{figure}

\paragraph{Patch-PnP} GDR-Net introduced a differentiable module named Patch-PnP, which is composed of convolutional layers followed by fully connected layers. This module processes geometric feature maps, such as 2D-3D Dense Correspondence Maps, and directly regresses the 6D object pose. Its primary aim is to replace the traditional PnP-RANSAC method with a more seamless and differentiable pipeline. The differentiable nature of Patch-PnP allows for end-to-end learning during the training of the pose estimation task.

\subsubsection{Multitask Pose Losses}
\label{sec:loss}

We utilize an ensemble of loss functions tailored to optimize different aspects of the 6D pose estimation task. The total loss comprises the following components:

\paragraph{Pose Loss} ($\mathcal{L}_\text{Pose}$) This loss is further disentangled into three components:
\begin{itemize}
    \item Rotation Loss ($\mathcal{L}_R$): Measures the discrepancy between the predicted and ground-truth rotations using a point matching loss. The rotation loss is computed as: 
    \begin{align*}
        \mathcal{L}_R = \frac{1}{N} \sum_{i=1}^{N} \| \hat{R} p_i - \bar{R} p_i \|_1
    \end{align*}
    where $\hat{R}$ is the estimated rotation matrix, $\bar{R}$ is the ground truth rotation matrix, $p_i$ are the 3D points in the object model, and $N$ is the total number of points.
    \item Center Loss ($\mathcal{L}_\text{center}$): $L1$ loss between the predicted and ground truth 2D center SITE components: 
    \begin{align*}
    \mathcal{L}_{\text{center}} = \| (\hat{\delta}_x, \hat{\delta}_y) - (\bar{\delta}_x, \bar{\delta}_y) \|_1 
    \end{align*}
    \item Depth Loss ($\mathcal{L}_z$): L1 loss between the predicted and ground truth object centroid SITE component: 
    \begin{align*}
    \mathcal{L}_z = \| \hat{\delta}_z - \bar{\delta}_z \|_1
    \end{align*}
\end{itemize}
\paragraph{Geometry Loss ($\mathcal{L}_\text{Geom}$)} This loss combines the dense correspondence loss and the mask losses:
\begin{itemize}
\item Mask Loss ($\mathcal{L}_\text{mask}$): The mask loss consists of two terms, the visible mask loss and the amodal (full) mask loss: 
\begin{align*}
\mathcal{L}_\text{mask} = \| \hat{M}_\text{vis} - \bar{M}_\text{vis} \|_1 + \| \hat{M}_\text{full} - \bar{M}_\text{full} \|_1
\end{align*}

where $M_\text{vis}$ is the visible object mask and $M_\text{full}$ is the full object mask.
\item 2D-3D Dense Correspondence Loss ($\mathcal{L}_\text{corr}$): $L1$ loss between the predicted and ground truth dense 2D-3D correspondence maps: 
\begin{align*}
    \mathcal{L}_{\text{corr}} = \frac{1}{\sum{\bar{M}_{\text{vis}}}} \| \bar{M}_{\text{vis}} \odot (\hat{D} - \bar{D}) \|_1
\end{align*}

where $D$ is the 2D-3D dense correspondence, and $\bar{M}_{\text{vis}}$ is the ground-truth visible object mask.
\end{itemize}
The total geometry loss is the sum of the dense correspondence loss and the mask loss: 
\begin{align*}
\mathcal{L}_\text{Geom} = \mathcal{L}_\text{corr} + \mathcal{L}_\text{mask}
\end{align*}
\paragraph{Articulation Loss ($L_\text{Art}$)} L1 loss between the predicted and ground truth articulation angles.
\begin{align*}
\mathcal{L}_z = \| \hat{A} - \bar{A} \|_1
\end{align*}
where A is the normalized articulation angle.
\paragraph{Object Category Loss ($L_\text{Cat}$)} Cross-entropy loss of the object categories.
\begin{align*}
\mathcal{L}_{\text{Cat}} = \text{CE}( \hat{\text{C}} , \bar{\text{C}} )
\end{align*}
where C is the is the object class.

The total loss is a weighted sum of these individual loss components:
\begin{equation}
    \mathcal{L}_{\text{total}} = w_{\text{Pose}} \cdot \mathcal{L}_{\text{Pose}} + w_{\text{Geom}} \cdot \mathcal{L}_{\text{Geom}} + w_{\text{Cat}} \cdot \mathcal{L}_{\text{Cat}} + w_{\text{Art}} \cdot \mathcal{L}_{\text{Art}}
\end{equation}

\subsubsection{Pose Model Architecture}
\label{architecture}
Our pose estimation network simultaneously estimates object category, articulation angle, and pose from image crops. The network consists of two interconnected sections: (1) the intermediate feature regression stage and (2) the pose regression stage (Figure \ref{fig:pose_model}).

The intermediate feature regression stage is based on UPerNet \cite{xiao2018unified}  with a Vision Transformer (ViT) backbone pretrained following the method described in \cite{oquab2023dinov2} on the LVD-142M dataset. The features extracted by the ViT backbone are fed into a Pyramid Pooling Module (PPM) \cite{zhao2017pyramid} before being fed back to the FPN. The object category head is attached to the PPM output for early branching \cite{elhoseiny2015convolutional}. The fused feature map from the FPN is directed to three heads: articulation, mask, and 2D-3D Dense Correspondence ($M_{2D-3D}$).

The pose regression stage varies depending on the phase. During training, the $M_{2D-3D}$ encoding is processed by Patch-PnP \cite{wang2021gdr}, a differentiable pose estimator. During testing, the $M_{2D-3D}$ encoding is converted to 2D-3D pairs, and the pose is estimated using a PnP-RANSAC variant. The final output consists of the pose parameters ($R_{3x3}, t_3$), object class, and articulation angle.

\subsubsection{Training Procedure}
\label{training_procedure}

The training involves iterative domain adaptation, starting with independent training on synthetic data. Pseudo-labels generated from one network’s predictions are used to retrain the other, and this process is repeated twice. Details are provided in the following subsections.

\paragraph{Step 1: Training on Synthetic Data} 
In the first step, the pose estimation network and the object detection network are trained independently on the synthetic dataset generated using the pipeline described in Section \ref{sec:synth}.

\paragraph{Step 2: Object Detection Domain Adaptation}
\begin{enumerate}
    \item Object tracking is applied to the real video data using the BoT-SORT tracking algorithm
    \item Frames with high detection confidence and at least three consecutive tracking detections are selected as pseudo-labels for object detection.
    \item The selected frames are processed by the pose estimator to predict the object pose, class, and articulation angle. The resulting mesh and pose are reprojected onto the image plane, and the reprojected bounding box is used as a new pseudo-label for the object detection network. This process is illustrated in Figure \ref{fig:bbox_ref}.
    \item The object detector is retrained on the pseudo-labeled data along with a fraction of the synthetic data.
\end{enumerate}

\paragraph{Step 3: Pose Estimation Domain Adaptation}
\label{domain_adapt}
\begin{enumerate}
    \item Image crops of the detected objects are obtained from the real video data using the refined object detection model.
    \item The pose estimation network is applied to these image crops to predict the 6D pose, object class, and articulation angle.
    \item Pose predictions are filtered based on class confidence, PnP-RANSAC outlier count, and reprojection error. Pose predictions that satisfy these filtering criteria are selected as pseudo-labels for retraining the pose estimation network. The specific numerical thresholds used for filtering are detailed in Section \ref{thresholds}. and a broader discussion of these criteria, along with potential future work, is provided in the Discussion section.
    \item The pose estimator is retrained on the pseudo-labeled data along with a fraction of synthetic data.
\end{enumerate}

\begin{figure*}[!t]
  \centering
  \includegraphics[scale=0.42]{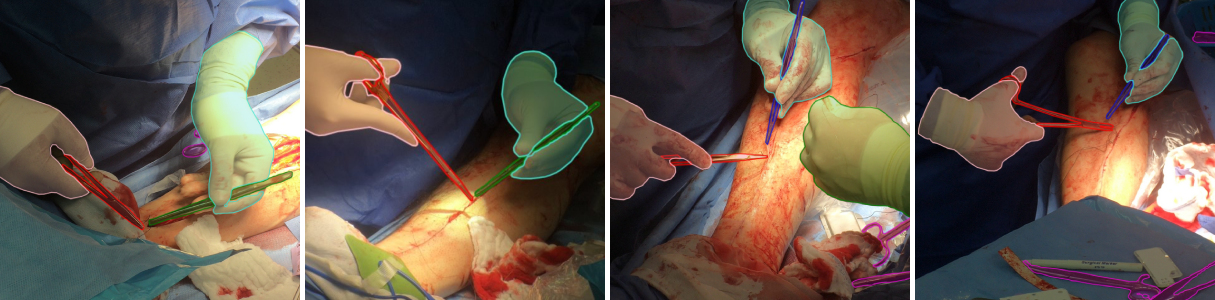}
  \caption{Annotated segmentation masks of surgical tools and hands.}
  \label{fig:hands_ab}
\end{figure*}

\section{Experiments, Results, and Ablation Studies}
\label{chap:fifthchap}

\subsection{Experimental Setup}
\label{exp_setup}
Our experiments are implemented with the PyTorch deep learning framework. We train our models end-to-end using the Ranger optimizer \cite{liu2019variance,zhang2019lookahead,yong2020gradient}, which combines Rectified Adam (RAdam) with Lookahead. A batch size of 120 is used during training, along with a base learning rate of 1e-5. We employ a cosine annealing learning rate schedule \cite{loshchilov2016sgdr}, which gradually reduces the learning rate after approximately 72\% of the training iterations.

During training, we apply various data augmentation techniques with a base probability of 0.8. These augmentations are taken from the GDRNPP implementation \cite{wang2021gdr, liu2022gdrnpp_bop} and include: Gaussian blur, sharpness enhancement, contrast enhancement, brightness adjustment, color enhancement, addition of random values to pixel intensities, channel-wise inversion, multiplication of pixel intensities, addition of Gaussian noise, linear contrast adjustment, and grayscale conversion.

For inference, we use the Progressive-X \cite{barath2019progressive} PnP-RANSAC scheme, which utilizes Graph-Cut RANSAC \cite{barath2018graph}.

\subsubsection{Dataset and Labels}
As discussed in Section \ref{sec:data_annot}, our real-world dataset consists of annotated frames from seven different surgical procedures. For each surgery, approximately 100 images were selected and manually annotated with detailed segmentation masks of the surgical tools and hands. These annotations are used for testing only; they serve as the ground truth for evaluating our pose estimation method on real-world data.

\subsubsection{Thresholds for Filtering Predictions}
\label{thresholds}
The thresholds used for filtering pose predictions during the domain adaptation step (Section \ref{domain_adapt}, Step 3) were chosen after evaluating their impact on the accuracy of pseudo-label generation during synthetic-to-real adaptation. The criteria are:
\begin{itemize}
    \item Class confidence: Predictions are retained if their class confidence score exceeds 0.999.
    \item PnP-RANSAC outlier count: Only predictions with an outlier count below 30 are accepted.
    \item Reprojection error: Predictions with a Jaccard index above 0.3 are selected.
\end{itemize}

\subsubsection{Evaluation Metrics}
Due to the lack of ground truth 6D pose annotations in our real-world dataset, we employ a 2D reprojection-based evaluation method \cite{brachmann2016uncertainty}. This approach compares the reprojected mask of the object, obtained using the estimated pose, class, and articulation angle, with the manually annotated segmentation masks. However, directly comparing the full reprojection masks to the annotated tool masks may not provide an accurate assessment of the pose estimation accuracy, as the annotated tool masks may be partially occluded by the surgeon's hands. To account for these occlusions, we incorporate the hand masks into our evaluation process.

\paragraph{6D Pose Evaluation Metric} We propose a hand-occlusion aware reprojection that focuses on evaluating the visible parts of the tools while considering the presence of hand occlusions. The steps involved in computing the metrics are as follows:
\begin{enumerate}
    \item Given an input image and the estimated pose of the surgical tool, we project the 3D model of the tool onto the 2D image plane using the camera's intrinsic parameters. This results in a reprojected binary mask representing the visible surface of the tool according to the estimated pose.
    \item We subtract the hand masks from both the annotated tool masks and the reprojected tool masks. This step removes the regions occluded by the hands, allowing us to focus on the visible parts of the tools.
    
    We use the Average Precision ($AP$) metric for segmentation following the method used in the COCO 2020 Object Segmentation Challenge \cite{lin2014microsoft} and BOP 2022 Challenge \cite{sundermeyer2023bop}. For each object, we calculate a per-object Average Precision ($AP_O$) by averaging the $AP$ across the following Intersection over Union (IoU) thresholds: [0.5, 0.55, 0.6, 0.65, 0.7, 0.75, 0.8, 0.85, 0.9, 0.95]. The IoU is evaluated between the manually subtracted annotated tool mask and the manually subtracted reprojected tool mask. The overall performance metric ($AP$) is then computed by averaging the $AP_O$ scores for all objects.
    
\end{enumerate}

\paragraph{2D Object Detection Metric} Similarly to our segmentation approach, the $AP$ is determined by calculating the precision at various IoU thresholds, yet here it specifically evaluates the accuracy with which objects are detected rather than segmented. The overall $AP$ score reflects the aggregate precision across all detected objects in the dataset.
We evaluate only those object instances where at least $10\%$ of their projected surface area is visible, excluding detections of objects visible from less than 10\% which are not counted towards false positives. Due to the lack of ground-truth poses and the minimal variation in camera-to-object distance, we estimate the visibility percentage of each object by counting the number of visible pixels in the ground truth segmentation.

\subsubsection{Results on Real-World Surgical Data}

\paragraph{Pose Estimation Results}
Table \ref{tab:pose_estimation_ap} presents the performance of our pose estimation method on the real-world surgical dataset. We evaluate the Average Precision (AP) for two surgical tools: needle-holder and tweezers. The AP is computed based on the hand-occlusion-aware reprojection metric described in the previous section.

We compare three variations of our method: (1) Pose estimation trained only on synthetic data, (2) Pose estimation with real data refinement. These variations are evaluated using both ground truth bounding boxes (GT Bbox) and predicted bounding boxes from the object detection stage.
It is important to note that predicting pose on the ground truth bounding boxes isolates the pose estimation performance and does not introduce the errors caused by the bounding box prediction.

\begin{table}[t!]
\caption{Pose Estimation Average Precision (AP) on the Real-World Surgical Data. "Synth" refers to training on synthetic data, "Real" refers to refinement on real data, "GT RoI" refers to ground truth RoI, and "Pred RoI" refers to predicted RoI.}
\centering
\begin{tabular}{ccccc}
\toprule
RoI & Trained on & Needle-holder & Tweezers & Mean \\
\midrule
GT RoI & Synth & 0.65 & 0.68 & 0.67 \\
GT RoI & Real & 0.77 & 0.82 & 0.79 \\
\midrule
Pred RoI & Real & 0.74 & 0.67 & 0.70 \\
\bottomrule
\end{tabular}
\label{tab:pose_estimation_ap}
\end{table}

\paragraph{Object Detection Results}
Table \ref{tab:object_detection_ap} presents the object detection performance on the real-world surgical dataset. We evaluate the Average Precision (AP) for detecting the needle-holder and tweezers using our object detection method.
We present results when training the model only on synthetic data, after refinement on real data and pose refinement with the tuned pose model as depicted in Figure \ref{fig:bbox_ref}.

\begin{table}[t!]
\caption{YOLOv8 Object Detection Average Precision on the Real-World Surgical Data}
\centering
\begin{tabular}{@{}lccc@{}}
\toprule
Training Strategy & Needle-holder & Tweezers & Mean \\
\midrule
trained on synthetic data & 0.56 & 0.45 & 0.51 \\
tuned on real data & 0.69 & 0.75 & 0.72 \\
pose refinement & \textbf{0.71} & \textbf{0.76} & \textbf{0.74} \\
\bottomrule
\end{tabular}
\label{tab:object_detection_ap}
\end{table}

\subsection{Ablation Studies}
\subsubsection{Impact of Synthetic Hand-Object Occlusions}
\label{ab:hands}
To investigate the impact of synthetic hand-object occlusions on the performance of our pose estimation and object detection models, we conduct a series of ablation studies. We train three variants of our models on different synthetic datasets. These datasets are a sub-sample of the whole synthetic data that we generated:
\begin{enumerate}
    \item Synthetic data with synthetic hands: This dataset includes realistic hand-object interactions, as described in Section \ref{sec:data_annot}.
    \item Synthetic data with masked tools: In this dataset, regions of the surgical tools that would have been occluded by hands were masked out using a rotating 2D bounding box that encapsulates the hand. This straightforward approach removes occluded tool regions without requiring complex hand modeling, avoiding potential biases while maintaining realistic tool occlusion patterns.
    \item Synthetic data without hands: Here, synthetic hands are excluded from the rendering process, leaving only the surgical tools visible in the scenes.
\end{enumerate}
Figure \ref{fig:hands_ab} shows an example of the hand dataset.
We evaluate the performance of the pose estimation model and the object detection model trained on each of these datasets.

\begin{figure*}[!t]
  \centering
  \includegraphics[scale=0.31]{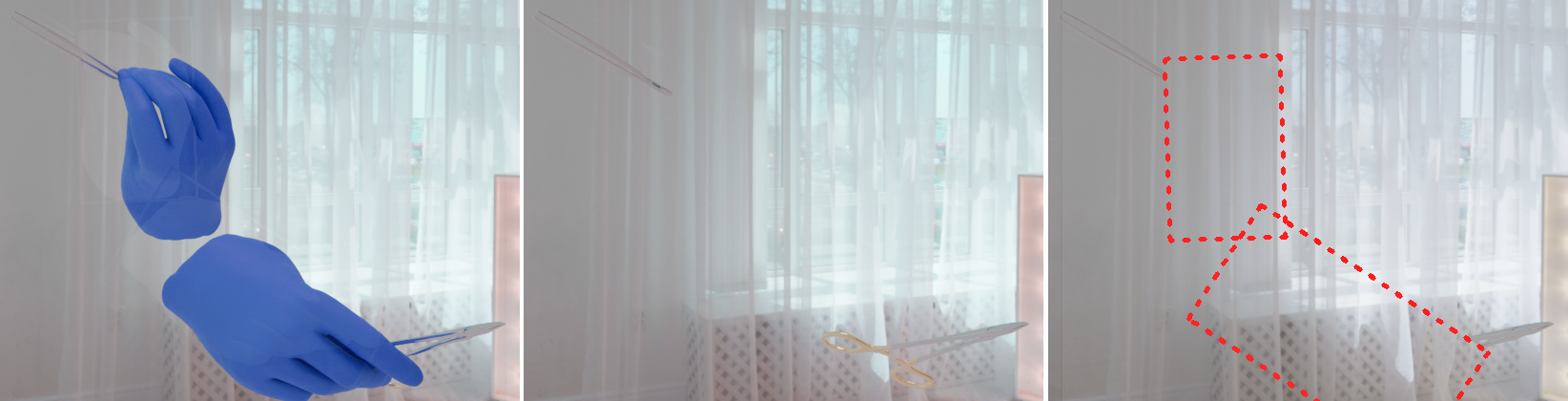}
  \caption{Examples from the synthetic hand ablation dataset illustrating occlusion variations used in Section \ref{ab:hands}. From left to right: image with synthetic hands, image without hands, and image with hands removed but tools masked at occluded regions. Rotated bounding boxes are overlaid to highlight the tool regions that are removed in the masking scenario. Additional examples are provided in the appendix.}
  \label{fig:hands_ab}
\end{figure*}

We evaluate the pose estimation accuracy of the models trained on the three synthetic datasets when provided with ground truth to isolate the impact of the object detection stage.

\begin{table}[h!]
\centering
\caption{AP Metrics for Pose Model Evaluation with Ground-Truth Bounding Boxes}
\begin{tabular}{@{}lccc@{}}
\toprule
Synthetic Dataset & Needle-holder & Tweezers & Mean \\ \midrule
Without Hands & 0.48           & 0.62      & 0.55 \\
Masked Tools     & 0.56           & 0.73      & 0.64 \\
With Hands        & \textbf{0.58}           & \textbf{0.76}      & \textbf{0.67} \\ \bottomrule
\end{tabular}
\label{eval-hands-pose}
\end{table}

\subsubsection{Patch-PnP vs. PnP-RANSAC}
\label{ab:pnp}
This ablation study evaluates the performance of two different strategies for pose estimation: Patch-PnP and PnP-RANSAC.

For PnP-RANSAC, we utilize the Progressive-x algorithm \cite{barath2019progressive} in conjunction with Graph-cut RANSAC \cite{barath2018graph}. Hereafter, we refer to this combination as PnP-RANSAC. Progressive-x is a multi-model fitting algorithm that progressively fits model instances to data points, taking into account both the fit quality and the spatial coherence of the points. Graph-cut RANSAC is a variant of RANSAC that enhances the efficiency and robustness of model fitting through graph-based segmentation. It employs a graph-based representation of the data points and uses graph-cut optimization to identify the optimal set of inliers for each model hypothesis.

\paragraph{Training strategy impact} We evaluate the impact of training an end-to-end pose estimation model with Patch-PnP and all the associated pose losses as described in section \ref{sec:loss}, vs training without the Patch-PnP module.
The inference strategy in both cases is with PnP-RANSAC. Results are shown in table \ref{tab:pnp_train_eval}.

\begin{table}[ht]
\caption{Training strategy comparison of Patch-PnP}
\centering
\begin{tabular}{cccc}
\toprule
Training Strategy & Needle-holder & Tweezers & Mean\\
\midrule
Without Patch-PnP & 0.57 & \textbf{0.80} & 0.68 \\
With Patch-PnP & \textbf{0.67} & 0.78 & \textbf{0.72} \\
\bottomrule
\end{tabular}
\label{tab:pnp_train_eval}
\end{table}

\paragraph{Inference strategy impact} 
In this experiment we evaluate PnP-RANSAC vs Patch-PnP inference on an end-to-end trained pose model. Results are shown in table \ref{tab:pnp_inf_eval}.

\begin{table}[ht]
\caption{Comparison of Patch-PnP and Prog-x for pose estimation inference.}
\centering
\begin{tabular}{cccc}
\toprule
Inference Strategy & Needle-holder & Tweezers & Mean \\
\midrule
Patch-PnP & 0.62 & \textbf{0.84} & 0.73 \\
Prog-x & \textbf{0.78} & 0.80 & \textbf{0.79} \\
\bottomrule
\end{tabular}
\label{tab:pnp_inf_eval}
\end{table}

\section{Discussion}
This study introduced a novel approach for monocular pose estimation of articulated surgical instruments in open surgery, marking an "in the wild" investigation in a relatively under-explored domain. Our approach involved creating a diverse synthetic dataset of surgical instruments with various articulation angles, enabling initial model training. These models were then refined through synthetic to real domain adaptation techniques, resulting in robust predictions of 6D pose, instrument class, and articulation angle in real surgical scenarios. The results demonstrate our method's effectiveness in addressing unique surgical environment challenges, such as limited real-world data, complex instrument geometry, and occlusions. Notably, our approach shows promise not only in pose estimation but also in improving object detection accuracy through an iterative refinement process that leverages our pose estimation method, suggesting a potential pathway for enhancing overall surgical instrument recognition systems.

\subsection{Results Overview}
Table \ref{tab:pose_estimation_ap} demonstrates the effectiveness of our approach in achieving robust pose estimation for real-world surgical tools. Our iterative domain adaptation pipeline improved the pose estimation Average Precision (AP) from 0.67 (synthetic-only training) to 0.79 when refined on real-world data with ground truth bounding boxes. When evaluated using predicted bounding boxes, the AP remains competitive at 0.70.

Similarly, object detection performance improved significantly through domain adaptation. As shown in Table 2, tuning on real data raised the detection AP from 0.51 to 0.72. Pose refinement further improved AP to 0.74.

\subsection{Occlusion Ablation Results}
Our occlusion ablation study (Section \ref{ab:hands}) highlighted the importance of incorporating hand-object interactions in the synthetic dataset. Training with realistic synthetic hand occlusions resulted in the highest pose estimation performance (mean AP: 0.67). These findings suggest that modeling hand occlusions is beneficial. However, we also found that simpler occlusion strategies, such as masking occluded regions without modeling hands, still yielded reasonable results, offering a practical alternative when detailed hand modeling is infeasible.

\subsection{Patch-PnP vs. PnP-RANSAC Ablation Results}
The ablation study comparing Patch-PnP and PnP-RANSAC (Section \ref{ab:pnp}) revealed insights into their respective strengths. When evaluating inference strategies, PnP-RANSAC achieved a higher mean pose estimation AP (0.79) compared to Patch-PnP (0.73), highlighting its superior performance in leveraging robust model fitting during testing. When used as a training module, the inclusion of Patch-PnP in an end-to-end pipeline resulted in improved model optimization, yielding a mean AP of 0.72 compared to 0.68 when training without it. These results suggest that while PnP-RANSAC excels in final inference robustness, Patch-PnP contributes to improving feature learning during training, making it a valuable component for network optimization.

\subsection{Benchmarking Against Prior Work}
Due to the unique nature of our task, directly benchmarking against prior work proves challenging. Although laparoscopic approaches primarily address cylindrical, minimally invasive tools, existing open surgery methods (e.g., \cite{hein2021towards,hein2023next}) focus on either joint hand and tool tracking or multi-view setups. These differences highlight the need for dedicated benchmarks and tailored methodologies in open surgery contexts. Although direct comparisons with these methods were not included in this study, future work that incorporates such comparative analyses would provide valuable insights into the relative strengths and limitations of our framework.

\section{Future Work}
While our approach shows promise, we acknowledge certain limitations that present opportunities for future research. Our current evaluation method focuses on the 2D reprojection of estimated poses, which does not fully capture the depth information of the instruments. Future work could incorporate depth information in the evaluation process to enhance 3D position accuracy. 

Additionally, our study was limited to two surgical
tools and one site, and expanding the synthetic dataset to include a wider variety of instruments would further demonstrate the generalizability of our approach.

Furthermore, our approach utilized several criteria to assess the quality of the pseudo-labels. Although these criteria effectively facilitated synthetic-to-real domain adaptation, a comprehensive hyperparameter search to evaluate their sensitivity and impact on downstream performance remains unexplored. Future work should investigate how these parameters influence the framework’s robustness across diverse domains.

\section{Conclusions}
In conclusion, this study demonstrates a novel approach to pose estimation of articulated surgical instruments in open surgery. Our approach involved creating a diverse synthetic dataset of surgical instruments with various articulation angles, enabling initial model training. These models were then refined through synthetic to real domain adaptation techniques, resulting in robust predictions of 6D pose, instrument class, and articulation angle in real surgical scenarios. The results demonstrate our method's effectiveness in addressing unique surgical environment challenges, such as limited real-world data, complex instrument geometry, and occlusions. Notably, our approach shows promise not only in pose estimation but also in improving object detection accuracy through an iterative refinement process.

This work contributes to ongoing efforts in computer-assisted surgery, offering a potential pathway to overcome the challenge of limited annotated real-world data. As research in this field progresses, we hope that this and similar approaches will continue to be refined, ultimately supporting the development of more effective tools for surgical assistance and training.

\bibliographystyle{elsarticle-num}\biboptions{numbers}
\bibliography{refs}

\end{document}